\def\year{2020}\relax
\documentclass[letterpaper]{article}
\usepackage{aaai20}
\usepackage{times}
\usepackage{helvet}
\usepackage{courier}
\usepackage{url}
\usepackage{graphicx}
\usepackage{booktabs}
\usepackage{amsmath, amssymb}
\usepackage{enumerate}
\usepackage{todonotes}
\usepackage{multirow}
\newcommand{\real}{\mathbb R}

\title{Who Did They Respond to? \\
Conversation Structure Modeling Using Masked Hierarchical Transformer}
\author{
Henghui Zhu,\textsuperscript{\rm 1}\thanks{Work done during an internship at the AWS AI Labs.}
Feng Nan,\textsuperscript{\rm 2}
Zhiguo Wang,\textsuperscript{\rm 2}
Ramesh Nallapati,\textsuperscript{\rm 2}
Bing Xiang\textsuperscript{\rm 2}\\
\textsuperscript{\rm 1}Boston University,
\textsuperscript{\rm 2}AWS AI Labs\\
henghuiz@bu.edu, \{nanfen, zhiguow, rnallapa, bxiang\}@amazon.com
}
\begin{document}
	
	\maketitle
	
	\begin{abstract}
        Conversation structure is useful for both understanding the nature of conversation dynamics and for providing features for many downstream applications such as summarization of conversations.
        In this work, we define the problem of conversation structure modeling as identifying the parent utterance(s) to which each utterance in the conversation responds to.
       
        Previous work usually took a pair of utterances to decide whether one utterance is the parent of the other. We believe the entire ancestral history is a very important information source to make accurate prediction. Therefore, we design a novel masking mechanism to guide the ancestor flow, and leverage the transformer model to aggregate all ancestors to predict parent utterances.
        Our experiments are performed on the {\it Reddit} dataset \cite{zhang2017characterizing} and the {\it Ubuntu IRC} dataset \cite{kummerfeld-etal-2019-large}. In addition, we also report experiments on a new larger corpus from the {\it Reddit} platform and release this dataset.
       
        We show that the proposed model, that takes into account the ancestral history of the conversation, significantly outperforms several strong baselines including the BERT model on all datasets.
    \end{abstract}
	
	\section{Introduction}
	
    When a group of people communicate with one another, there exist inherent structures in the conversation. One of the structures can be defined as the `reply\_to' relationship between a pair of utterances. The `reply\_to' relationship may be explicitly defined in some platforms including Reddit, Twitter, and Facebook. In other platforms such as Internet Relay Chat (IRC), Slack, and most other forums, there is no explicit `reply\_to' relationship in the conversation.     
    The problem also exists in Automatic Speech Recognition (ASR) transcripts of recorded conversations where the output typically consists of a flat list of utterances with no structure assigned to them. Identification of the structure of such conversations typically entails significant human labeling effort \cite{kummerfeld-etal-2019-large}.
    
    The problem of discovering conversation structure is also referred to as conversation disentanglement in the literature \cite{elsner-charniak-2008-talking,kummerfeld-etal-2019-large}. Disentangling the conversation structure opens the door for many downstream applications. For example, the topics discussed in a conversation are often non-contiguous and intertwined with each other. Discovering the conversation structure allows us to segment the conversation by topics. It also permits easier analysis and summarization of the various threads and topics discussed in the conversation. In addition, conversation structure can also be used to improve discourse act classification, which is useful in dialogue modeling \cite{zhang2017characterizing}.
    \begin{figure}[htbp]
    	\includegraphics[width=\columnwidth]{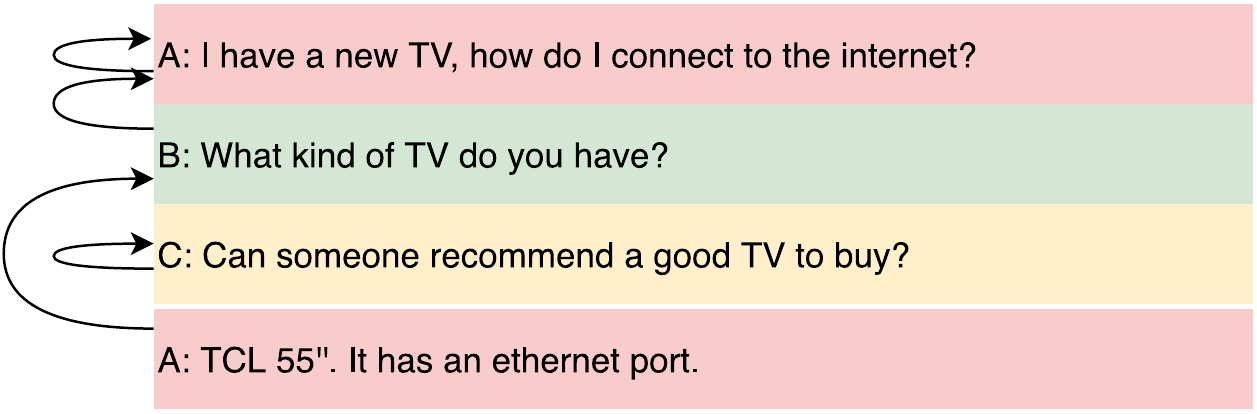}
    	\caption{An example sequence of utterances. Curved lines with arrows are the ground truth `reply\_to' relationship in the conversation. Without the context of the first utterance, it is hard to tell whether the fourth utterance is responding to the second utterance or the third utterance.}
    	\label{fig:example_conversation2}
    \end{figure}

    Previous work \cite{kummerfeld-etal-2019-large} only takes a pair of utterances to decide whether one utterance is the parent of the other without considering the context. Consider an example conversation shown in Figure \ref{fig:example_conversation2}. User A poses a question in the first line; user B asks a clarification question in the second line responding to the first line; user C then interjects with an unrelated question in the third line; finally, user A makes a statement in the fourth line.  Judging in a pairwise fashion, without considering context, the last line is equally likely to be replying to the questions in the second and third lines. However, by considering the context of the first utterance, it is clear that user A is continuing a conversation with user B to address the internet connection question.

    The goal of this work is to automatically discover the structure of a conversation given all the utterances. The task is very challenging since (a) the `reply\_to' relationship can exhibit long-distance behavior, i.e., an utterance may response to an utterance that occurred several turns earlier in the conversation, and (b) various topics can be intertwined with one another in the conversation. 
    
    In this work, we propose a new {\it Masked Hierarchical Transformer} model to learn conversation structures taking into account the prior context of each utterance. Concretely, a pre-trained transformer is used to produce a feature vector of each utterance. Then a second-stage transformer is applied to compare and aggregate utterances and produce the probability for each utterance being the parent of a target utterance. To utilize the previous conversation structure information, we introduce a novel masking mechanism for the second-stage transformer. We evaluate our model on three datasets using metrics for accurately identifying the `reply\_to' edges as well as metrics for reconstructing the whole conversation structure. We demonstrate that our method outperforms several strong baseline models including a BERT-based sentence-pair classification model, and achieves state-of-art results on all three conversation structure modeling datasets.

	\section{Related Work}
	\paragraph{Next Utterance Prediction} Next-utterance prediction is very related to our task of conversation structure modeling. There are two types of formulation of next-utterance prediction. The first one is to generate the next utterance in a conversation given the conversation history \cite{zhao2019effective,dziri2018augmenting,hu2019gsn} and the second type is to retrieve the next utterance in the conversation from a large list of response utterances \cite{lowe-etal-2015-ubuntu,whang2019domain}. These tasks are useful for building a chatbot, aiming to generate or retrieve a good response according to the context of the conversation, while this work aims to recover the structure of the entire conversation.
	
	\paragraph{Conversation Disentanglement} Conversation structure modeling task is also related to the conversation disentanglement task \cite{Shen2016detect,elsner-charniak-2008-talking,kummerfeld-etal-2019-large}. In particular, \cite{kummerfeld-etal-2019-large} proposed a large corpus for conversation disentanglement and a feed-forward model using average word embedding of sentences and some additional features. Conversation disentanglement is one type of conversation structure modeling task, where several conversation threads are interleaved and need to be disentangled. The ground truth `reply\_to' labels need to be annotated manually. In this paper, we have included the conversation disentanglement task for Ubuntu IRC in \cite{kummerfeld-etal-2019-large} into our analysis as well and compare with their model.
	
	\paragraph{Discourse Parsing} Conversational discourse parsing is also related to this task \cite{Joty:2012:NDF:2390948.2391047,afantenos-etal-2015-discourse,AAAI1714800}. Discourse parsing usually considers different kinds of relationship such as `question-answer' and 'acknowledgment'. The focus of these papers is mostly on the classification of the type of relationship, which is not the focus of this work.
	
	\section{Model}
	
	Consider the list of utterances in a conversation denoted as $S_1,\, S_2\cdots,\,S_N$. We aim to find out which of the utterance(s) in $S_1,\, S_2\cdots,\,S_L$ is the parent utterance of the target utterance $S_L$ for $L \in \{2,\cdots N\}$. (In Ubuntu IRC dataset, the parent of an utterance can be itself.)	We may access the structure of the conversation history during training (or its predicted structure at test time) i.e., all the parent(s) of $S_i$ are assumed to be known when we are predicting whether $S_i$ is the parent of the target utterance $S_L$, where $i < L$ and $L \in \{2,\cdots N\}$. 
	
	\subsection{Baseline Models} 
	The conversation structure modeling task can be viewed as an utterance pair classification task for the `reply\_to' relationship. Therefore, several existing methods for sentence-pair classification can be applied to this problem. In this paper, we consider the following models: Decomposable Attention Model \cite{parikh-etal-2016-decomposable}, Enhanced Sequential Inference Model (ESIM) \cite{chen-etal-2018-neural-natural} as well as the state-of-the-art BERT model \cite{devlin-etal-2019-bert} for sentence pair classification.  The decomposable attention model uses an attention mechanism that compares and aggregates the token-level information in both sentences to produce a set of meaningful features for sentence pair classification. ESIM improves this model by adding two LSTMs before and after the attention for including more contextual information in the feature vectors. BERT uses a transformer model \cite{vaswani2017attention} to perform comparison and aggregation, and leverages the unsupervised pretraining on a large corpus. In this paper, we use Glove embedding \cite{pennington2014glove} in the decomposable attention model and ESIM. For Reddit-small dataset, we also consider ELMo \cite{peters-etal-2018-deep} as word embedding for comparison. Also, since there are usually more non-parent-relation sentence pairs compared with parent-relation pairs, this paper balances the dataset by downsampling the non-parent sentence pairs when training.
	
	\subsection{Masked Hierarchical Transformer Model}
	
\begin{figure}[htbp]
		\includegraphics[width=\columnwidth]{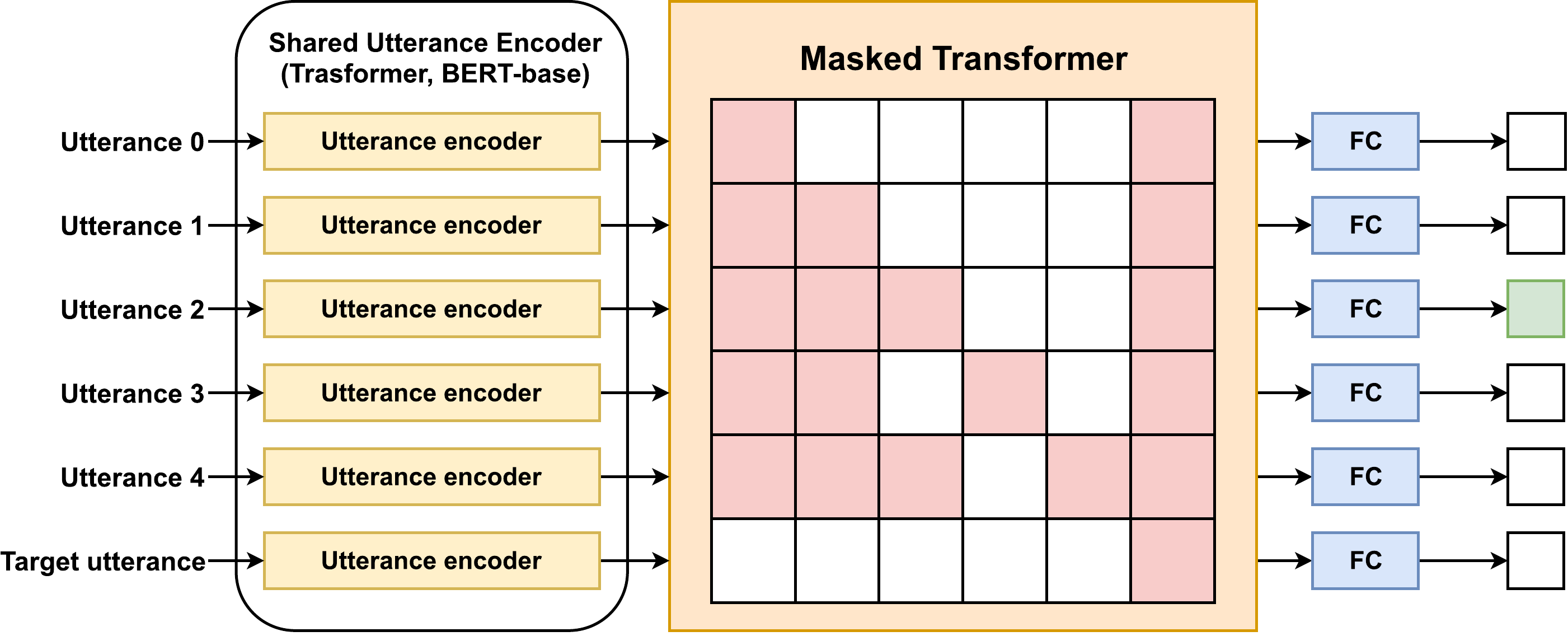}
		\caption{Diagram of the masked hierarchical transformer for conversation structure modeling. The colored blocks on the right indicates one element in the mask matrix, which means the corresponding utterance is attendable. The white block, on the other hand, indicates a zero element.}
		\label{fig:diagram}
	\end{figure}

    The baseline methods take into account only two utterances in the conversation, not the context of the conversation. Also, the structure of the conversation history may help in identifying the parent utterance(s) for a target utterance. Therefore, we consider a Masked Hierarchical Transformer model for the conversation modeling task, displayed in Figure \ref{fig:diagram}. The input of the model is a sequence of history utterances $S_1,\, S_2\cdots,\,S_L$, where the target utterance is placed at the end of the sequence as $S_L$. For each utterance, a shared utterance encoder (the yellow boxes in Figure \ref{fig:diagram}) is first used to produce a feature vector. In particular, we use a transformer model with the same configuration as \texttt{BERT-base, uncased} \cite{devlin-etal-2019-bert}, initialized with the pre-trained model. The output of the reserved [CLS] token in the BERT model is used as the feature vector of each utterance, denoted as $V_1,\, V_2\cdots,\,V_L$. Then we apply a second-stage masked transformer (the orange boxes in Figure \ref{fig:diagram}) to compare and aggregate information between the target utterance and each of the parent utterance candidates. We use a 4-layer transformer, each layer is with hidden size 300, intermediate states size 1024, and 4 attention heads. In order to use conversation structure, rather than the entire history, we design a masking mechanism to filter out redundant utterances, and only leverage ancestral utterances to predict the target utterance. The details of the masking will be discussed in the next subsection. Then, we denote the output of the second-stage transformer for each utterance as $\tilde{V}_1,\, \tilde{V}_2\cdots,\,\tilde{V}_L$. Finally, a fully-connected layer is used to produce the logits for the classification problem. For the Reddit datasets, since one comment has only one parent comment, we apply a softmax after the fully-connected layer and use a ranking loss. In particular,
	\begin{align*}
	t_i &= W_o \tilde{V}_i + b_o,\, i=1, \cdots, L-1, \\
	\hat{Y}_i &= \frac{\exp(t_i)}{\sum_{j=1}^{L-1} \exp(t_j)},\, i=1, \cdots, L-1,
	\end{align*}
	and the rank loss is
	\begin{equation}
	- \sum_{i=1}^{L-1} y_i \log(\hat{Y}_i), \label{eq:rank_loss}
	\end{equation}
	where $W_o \in \real^{1 \times 300}$ and $b_o \in \real$ are the parameters in the last fully-connected layer and $y_i$ is the binary indicator for if utterance $i$ is the parent utterance of the target utterance.
	
	For the Ubuntu IRC dataset, one utterance may have more than one parent. Also, the parent of an utterance can be itself. Therefore, we use a binary cross-entroy loss for each comment. In particular, the loss is defined as 
	\begin{gather}
	- \sum_{i=1}^{L} y_i \log\left( \sigma\left(t_i \right) \right) + (1-y_i) \log\left(1- \sigma\left(t_i\right) \right), \label{eq:sigmoid} \\
	t_i = W_o \tilde{V}_i + b_o,\, i=1, \cdots, L, \notag
	\end{gather}
	where $W_o \in \real^{1 \times 300}$ and $b_o \in \real$ are again the parameters in the last fully-connected layer and $y_i$ is the binary indicator for if utterance $i$ is the parent utterance of the target utterance.
	
	\paragraph{Ancestor Masking}
	
    The mask $\mathbf{M}$ in the second-stage transformer (the orange boxes in Figure \ref{fig:diagram}) is an asymmetric binary matrix of size $L\times L$ that encodes whether or not utterance $i$ attends to utterance $j$ ($\mathbf{M}_{ij}=1$) or not ($\mathbf{M}_{ij}=0$). Instead of leveraging all history, we design the masking mechanism to aggregate information only from an utterance's ancestors. In particular, the following properties should hold true: 
    \begin{enumerate}[1)]
    	\item All history utterances are able to attend to the target utterance, since all history utterances are parent candidates for the target utterance. And the transformer aim to produce features of the utterance being the parent of the target utterance.
    	\item Every utterance is able to attend to itself. 
    	\item Every non-target utterance is able to attend to its ancestors in the conversation graph.
    	\item All remaining utterances should not be attended to.
    \end{enumerate}

\begin{figure}[htbp]
		\includegraphics[width=\columnwidth]{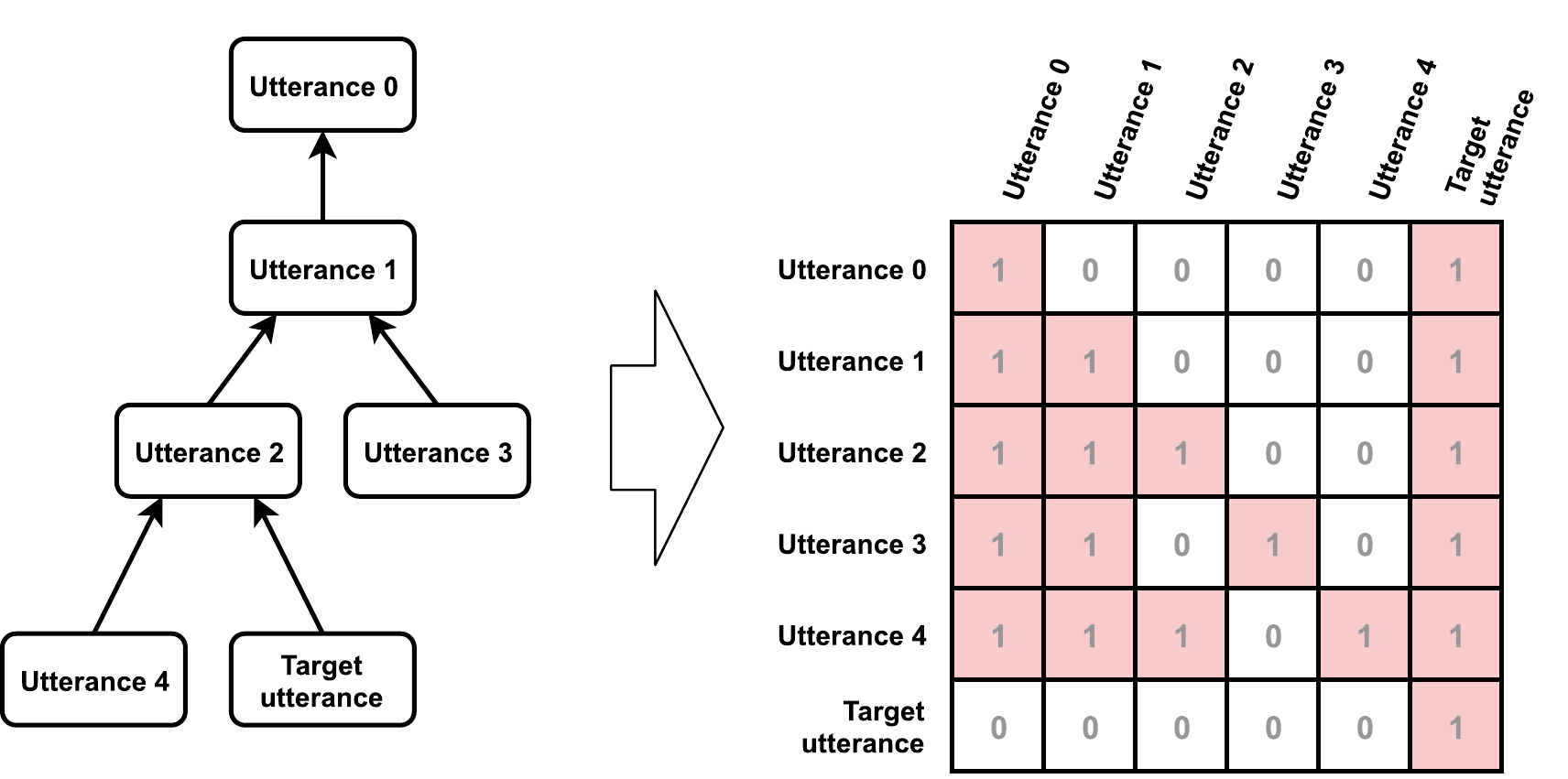}
		\caption{Graphical representation of the mask matrix for the second-stage transformer.}
		\label{fig:mask}
	\end{figure}

    Consider a conversation example on the left side of Figure \ref{fig:mask}. The mask matrix is shown on the right, denoted as $\mathbf{M}$. The last column of the matrix is all one due to property (1) above. The diagonal elements of the matrix are one due to property (2). Finally, for property (3), consider the following example pairs. Utterance 1 is the ancestor for utterance 4. Therefore, utterance 1 can be attended to by utterance 4 and the corresponding element in the mask matrix $\mathbf{M}_{41} = 1$. On the other hand, utterance 3 is not the ancestor of utterance 4. Therefore, the information of utterance 3 is not accessible to utterance 4 and $\mathbf{M}_{43} = 0$.
	
    For each target utterance, we generate this asymmetric mask matrix that encodes the conversation structure until that point in time and run the top transformer on the entire sequence of utterances seen thus far. At training time, the mask is defined based on the ground-truth conversation structure, while at test time, we define the mask based on previous predictions.
	
    \paragraph{Two-Stage Training} The shared utterance encoder in the masked hierarchical transformer model is initialized with a pre-trained BERT model, while there is no pre-training for the second-stage transformer initially. This may cause issues during model training: if one uses a large learning rate of the optimizer during training, the parameters in the shared utterance encoder may be washed out, causing what is known as catastrophic forgetting. But if one applies a small learning rate, it may take a very long time for the top transformer to converge. To address this problem, this paper proposes a two-stage training approach. In the first stage, we freeze the parameters of the shared utterance encoder after initialization with the pre-trained model during the training. In this case, we are essentially training the second-stage transformer and a relatively large learning rate can be applied. After this stage, the parameters in the shared utterance encoder are unfrozen and all the parameters in the model are jointly trained with a relatively small learning rate.
	
	\section{Datasets}
	
	\paragraph{Reddit-Small} 
	
    The comments in Reddit are organized by the `reply\_to' relationship, and the relationship graph is a tree whose root node is the title comment. The first dataset we consider is the selection of the Reddit conversations\footnote{We define a conversation as a submission in Reddit. Each comment in Reddit is equivalent to an utterance in a conversation.} from \cite{zhang2017characterizing}\footnote{They only provide the comment IDs and crawling scripts, the data generated in this paper is crawled on 05/24/2019. Also, the comments and conversation in Reddit change every day, i.e., an existing comment can be deleted. Therefore, there are many comments missing when we collected the data.}. We call this dataset {\it Reddit-small} in this paper. This dataset has 9,483 conversations. The following rule is used for filtering the data: if a comment is deleted, the comment itself and all its descendants are considered invalid. Then we split the conversations into train/dev/test sets using an 80/10/10 ratio. Also, we include only the first 16 comments in a Reddit conversation in the test set due to the capacity of the model. We find very few conversations have more than 16 comments in this selection. The statistics of three sets are shown in Table \ref{tab:stat_reddit}. The maximum path length of the conversation tree in those Reddit conversations is less than 5, which indicates the tree structures of the conversations are not very deep.
	
	\begin{table}[htbp]
		\centering
		\begin{tabular}{cccc}
			\toprule
			Split & \# comments & \# conversations & Average depth\\
			\midrule
			Train & 88,754 & 7,487 & 4.49\\
			Dev & 11,056 & 936 & 4.43\\
			Test & 11,524 & 936 & 4.61\\
			\bottomrule
		\end{tabular}
		\caption{Statistics of the Reddit-small dataset.}
		\label{tab:stat_reddit}
	\end{table}
	
	\paragraph{Reddit-Large}
	
  Since the conversation tree structure in Reddit-small dataset is relatively flat, In this paper, we create a relatively challenging dataset from Reddit dump\footnote{\url{https://files.pushshift.io/reddit/}}. We filter the {\it sub-reddits} to remove those that are  \texttt{Over\_18} or on the Not Safe For Work (NFWS) list\footnote{\url{https://www.reddit.com/r/NSFW411/wiki/index}}. The comments that have non-ASCII characters, deleted, or contain more than 128 characters are dropped. If one comment is dropped, so will be all of its decedents. Next, we remove the conversations that have a maximum tree depth less than 6. Finally, we randomly sample 10\% of the conversations that satisfy the above conditions. We call this dataset {\it Reddit-large} in this paper.\footnote{We release the dataset at \url{https://github.com/henghuiz/MaskedHierarchicalTransformer}} Since we observe the conversations in Reddit-large are longer compared with the Reddit-small, a same algorithm usually has a lower performance in this dataset. To make the evaluation result comparable to the previous Reddit dataset, we randomly remove the leaves in the conversation tree till the number of the comments in the conversation is 16, if there are more than 16 comments in the conversation. The statistics of train/dev/test set is shown in Table \ref{tab:stat_reddit_dump}. Compared with the Reddit-small dataset, the data is 30x larger and the conversation trees are deeper, which makes reconstruction of the conversation more challenging. 
	\begin{table}[htbp]
		\centering
		\begin{tabular}{cccc}
			\toprule
			Split & \# comments & \# conversations & Average depth\\
			\midrule
			Train & 2,872,524 & 57,196 & 7.12\\
			Dev & 315,058 & 6,356 & 7.16\\
			Test & 202,226 & 12,638  & 6.82\\
			\bottomrule
		\end{tabular}
		\caption{Statistics of the Reddit-large dataset.}
		\label{tab:stat_reddit_dump}
	\end{table}
	
	\paragraph{Ubuntu IRC}
	
    Finally, we consider the Ubuntu IRC dataset \cite{kummerfeld-etal-2019-large}. It consists of messages manually annotated with `reply\_to' structures that disentangle conversations.
   
    This work only considers the training and development portion of the data\footnote{The test set annotation was not released when submitting this paper.}, where we use the official dev set in \cite{kummerfeld-etal-2019-large} as the test set in our experiments. We randomly select 140 conversations\footnote{ In Ubuntu IRC dataset, we define a conversation as a sample of continuous conversation history in the channel, usually consisting of 100-500 messages and 1000 additional context messages before those messages considered.} of the train set in their paper as our training set, and the remaining 13 conversations are used as our development set. The statistics of the dataset used in this paper are shown in Table \ref{tab:stat_irc}. Compared with the Reddit dataset, an utterance in IRC may have more than one parent from the conversation history, although 97.40\% of utterances have only one parent. Furthermore, for an utterance in Ubuntu IRC, the parent utterance may be itself. Finally, there are occurrences of system-generated messages in this corpus along with human-generated messages.
	
	\begin{table}[htbp]
		\centering
		\begin{tabular}{cccc}
			\toprule
			Split & \# conversation & \# ann. messages & Avg. \# parent
			\\
			\midrule
			Train & 140 & 61561 & 1.03\\
			Dev & 13 & 5902 & 1.03\\
			Test & 10 & 2500 & 1.04\\
			\bottomrule
		\end{tabular}
		\caption{Statistics of the Ubuntu IRC dataset. (The test set in this paper is the dev set in \cite{kummerfeld-etal-2019-large})}
		\label{tab:stat_irc}
	\end{table}
	
	\section{Experimental Results}
	
	\subsection{Model Performance on Reddit datasets}
	
	We first consider the Reddit-small dataset. In this dataset, each comment has a unique parent comment, which excludes itself. Therefore, we use a rank loss defined in Equation \eqref{eq:rank_loss} in the masked hierarchical model.
	
	Since the number of comments in a Reddit thread is relatively small but the length of comments is long, we limit the length of each comment to be 50 subwords after byte-pair-encoding (BPE) tokenization (51st word and on-wards are removed). Also, we limit the length of conversation history to be 16 during training by iteratively removing leaf nodes in the conversation graph so that the target comment and the title comment (root node) are in the conversation history. The input to our model is the comment history sorted by the timestamps of the utterances. 
	
    We consider two metrics to evaluate the performance. The \emph{graph accuracy} is defined as the average agreement between the ground truth and predicted parent for each utterance. The \emph{conversation accuracy} is defined as the agreement between the entire conversation tree structure reconstructed by the model and the ground-truth structure.
	
    All baseline models are trained with Adam optimizer \cite{kingma2014adam}. For the decomposable attention model and the ESIM, we use a learning rate $10^{-4}$ with batch size 32 for 50 epochs. For the BERT model, we use a base learning rate $3 \times 10^{-5}$ for 10 epochs with the same learning schedule describe in \cite{devlin-etal-2019-bert}.    For training our model, we use a learning rate $10^{-4}$ for pre-training the second-stage transformer with batch size 32 for 10 epochs. Then we use a learning rate $10^{-5}$ and batch size 8 for training all the parameters of the model for 10 epochs again. The smaller batch size for the second stage of training is due to the larger model capacity during training. We apply early stopping for both baseline models and the two training stages of our approaches. When training, the ground truth conversation tree structure is used to generate the mask. During evaluation phase, we use the predicted conversation tree structure by the model to generate the mask.
	
    Table \ref{tab:performance_mit} shows the results of the baseline models and our model in the Reddit-small dataset. We also include a naive method called \texttt{Predict first} as a baseline, that simply returns the title comment in the Reddit conversation. Also, we include ELMo\footnote{This paper uses ELMo embedding from Tensorflow Hubs \url{https://tfhub.dev/google/elmo/2}.} as an alternative for the word embedding for the two embedding-based baselines, namely the decomposable attention model and the ESIM model. Finally, we also use the BERT model that makes pairwise decisions, as a state-of-the-art baseline. As shown in Table \ref{tab:performance_mit}, \texttt{Predict first} is a strong baseline due to the fact the conversation trees are relatively flat in Reddit-small dataset (see Table \ref{tab:stat_reddit}). Adding ELMo embedding improve the model accuracy. We run our masked hierarchical transformer model with 5 initial random seeds and report the average and the standard deviation of the score. As shown in the table, our novel approach outperforms the baseline models by a large margin in both graph accuracy and conversation accuracy.
	
	\begin{table}[htbp]
		\centering
		\begin{tabular}{ccc}
			\toprule
			Method & Graph Acc. & Conv. Acc. \\
			\midrule
			Predict first & 45.60\% & 15.28\%\\
			\midrule
			Decom. Atten. w/ Glove & 43.08\% & 12.22\%\\
			Decom. Atten. w/ ELMo & 45.82\% & 12.92\%\\
			ESIM w/ Glove & 42.92\% & 12.08\%\\
			ESIM w/ ELMo & 48.43\% & 14.31\%\\
			BERT & 55.86\% & 17.78\%\\
			\midrule
			\multirow{2}{*}{Our Approach} & \textbf{60.53\%} & \textbf{20.61\%} \\
			& (0.34\%) & (0.47\%)\\
			\bottomrule
		\end{tabular}
		\caption{Performance of the models in Reddit-small dataset. The numbers in the parentheses are the standard deviation of the score over 5 runs.}
		\label{tab:performance_mit}
	\end{table}
	
	Shown in Table \ref{tab:performance_dump} is the performance comparison of all the models on Reddit-large dataset. Again, we see that our approach outperforms the baselines by a large margin.
	
	\begin{table}[htbp]
		\centering
		\begin{tabular}{ccc}
			\toprule
			Method & Graph Acc. & Conv. Acc. \\
			\midrule 
			Predict first & 33.60\% & 0.06\% \\
			\midrule
			Decom. Atten. w/ Glove & 24.86\% & 0.02\%\\
			ESIM w/ Glove & 18.79\% & 0.00\%\\
			BERT & 39.48\% & 0.04\%\\
			\midrule
			Our Approach & \textbf{42.87\%} & \textbf{0.13\%} \\
			\bottomrule
		\end{tabular}
		\caption{Performance of the models in Reddit-large dataset}
		\label{tab:performance_dump}
	\end{table}
	\subsection{Model Performance on Ubuntu IRC dataset}
	
	Compared with the Reddit datasets, an utterance in the Ubuntu IRC dataset may have more than one parent. Therefore, we consider the binary cross-entropy loss defined in Equation \eqref{eq:sigmoid} for the Ubuntu IRC disentanglement problem. 
	
	For training the masked hierarchical transformer model, we include the most 40 recent messages including the target message to determine the parent message(s). Since the messages are usually short, we limit their length to 36 subwords after tokenization. In this dataset, there are no parent message annotations for the first 1000 messages of each part, known as the context messages. Therefore, we defined the messages itself as its parent for all the context messages.
	
	Similar to the Reddit datasets, we use Adam optimizer with a learning rate $10^{-5}$ for pre-training the top transformer with batch size 32 for 10 epochs. Then we use a learning $10^{-7}$ and batch size 4 for training all the parameters of the model for 10 epochs again. Compared with the training configuration used for the Reddit datasets, it uses a smaller learning rate and batch size due to larger model size during computation and fewer samples. Again, early stopping is used for these two training stages. Also, since most of the utterances in this dataset have only one parent, we consider the utterance in the conversation history that has the largest probability to be the parent of a target utterance during test time.
	
  As shown in \cite{kummerfeld-etal-2019-large}, feature-based models work well on this dataset. The features consist of a few global-level features including year and frequency of the conversation, utterance level features including types of message, targeted or not, time difference between the last message, etc., and utterance pair features including how far apart in position and time between the messages, whether one message targets another, etc. Since these features have proved to be very useful in predicting the parent relationship in the Ubuntu IRC dataset, we also consider concatenating them with the utterance features vector before feeding to the second-stage transformer.
	
  Graph metrics for Ubunut IRC dataset are first considered including the precision, recall and F1 scores for the parent relationship prediction. We first consider the performance of the baseline models without any additional features. For the decomposable attention model and ESIM, we use the Glove embedding by \cite{kummerfeld-etal-2019-large}. Our approach, shown as the second group in Table \ref{tab:ubuntu-graph}, outperforms the baseline models significantly. Besides, we consider models that use those features in \cite{kummerfeld-etal-2019-large}. Since our test set is the development set in their paper and no results have been reported, we train the \texttt{Linear} and \texttt{FeedForward} model in their paper under our train/dev split and evaluate them on our test set. Also, we include those features into both baseline models and our model in the results shown as the third group in Table \ref{tab:ubuntu-graph}. Again, the performance of our model is the best among all those models.
	
	\begin{table}[htbp]
		\centering
		\begin{tabular}{cccc}
			\toprule
			Method & P  & R  & F\\
			\midrule
			Previous & 30.8 & 29.5 & 30.2\\
			Linear * \cite{kummerfeld-etal-2019-large} & 64.2 & 61.6 & 62.9\\
			FF * \cite{kummerfeld-etal-2019-large} & 69.3 & 66.5 & 67.9 \\
			\midrule
			Decom. Atten. w/ Glove & 27.6 & 26.6 & 27.1\\
			ESIM w/ Glove & 28.7 & 27.8 & 28.2\\
			BERT & 34.8 & 33.7 & 34.2\\
			Our Approach  & 53.9 & 51.7 & 52.8\\
			\midrule
			Decom. Atten. w/ Glove + F * & 65.8 & 63.1 & 64.4\\
			ESIM w/ Glove + F * & 64.7 & 61.9 & 63.3\\
			BERT + F * & 67.6 & 64.9 & 66.2\\
			Our Approach + F * & \textbf{73.2} & \textbf{69.2} & \textbf{70.6}\\
			\bottomrule
		\end{tabular}
		\caption{Graph results on the Ubuntu IRC test set (Dev set in \cite{kummerfeld-etal-2019-large}). The methods with suffix `*' are using the additional hand-crafted features in \cite{kummerfeld-etal-2019-large}. The baseline models with `+F' suffix are with the additional features.}
		\label{tab:ubuntu-graph}
	\end{table}
	
	Table \ref{tab:ubuntu-conversation} shows the conversation results of the baseline methods and our proposed one using the same evaluation metrics as \cite{kummerfeld-etal-2019-large}. In particular, a cluster in a conversation is defined as a connected component in the conversation graph. Three types of metrics are considered as the modified Variation of Information (VI) in \cite{kummerfeld-etal-2019-large}, One-to-One Overlap (1-1) of the cluster \cite{elsner-charniak-2008-talking}, and the precision, recall, and F1 score between the cluster prediction and ground truth. Our approach combined with the additional features achieves the best performance among all the metrics. This suggests our model is both good at finding the correct parent of an utterance and reconstructing the whole conversation.
	
	\begin{table}[htbp]
		\centering
		\begin{tabular}{cccccc}
			\toprule
			Method & VI & 1-1 & P  & R  & F\\
			\midrule
			Previous & 66.2 & 23.7 & 0.0 & 0.0 & 0.0\\
			Linear *  & 87.5 & 66.8 & 17.8 & 26.0 & 21.1\\
			FF *  & 88.8 & 71.3 & 28.0 & \textbf{32.7} & 30.2\\
			\midrule
			Decom. Atten & 70.3 & 39.8 & 0.6 & 0.9 & 0.7\\
			ESIM & 72.1 & 44.0 & 1.4 & 2.2 & 1.8\\
			BERT & 74.7 & 45.4 & 2.2 & 3.6 & 2.7\\
			Our Approach  & 82.1 & 59.6 & 8.7 & 12.6 & 10.3\\
			\midrule
			Decom. Atten. + F * & 87.4 & 66.6 & 18.2 & 25.1 & 21.1\\
			ESIM + F * & 87.7 & 65.8 & 18.9 & 28.3 & 22.6\\
			BERT + F * & 89.5 & 71.7 & 21.4 & 30.0 & 25.0\\
			Our Approach + F * & \textbf{89.8} & \textbf{75.4} & \textbf{35.8} & \textbf{32.7} & \textbf{34.2}\\
			\bottomrule
		\end{tabular}
		\caption{Conversation results on the Ubuntu IRC test set (dev set in \cite{kummerfeld-etal-2019-large}). The methods with suffix `*' are using the additional hand-crafted features in \cite{kummerfeld-etal-2019-large}. The baseline models with `+F' suffix are with the additional features.}
		\label{tab:ubuntu-conversation}
	\end{table}
	
	\section{Ablation Studies}
	
	\paragraph{Importance of the Mask} 
	
    The ablation study is performed on Reddit-small dataset. We aim to show the importance of the mask used in our model. First, we consider training an alternative model of the hierarchical transformer model without masking. In this case, the utterance is able to attend to all the utterances. Intuitively, it is hard for this model to come up with a good prediction, since attention may be distracted by all the utterances. The results in Table \ref{tab:ablation_mask} show a huge drop of performance when the no mask is used, indicating that the ancestor history plays a very important role in recovering the structure of the conversation.
	
	\begin{table}[htbp]
		\centering
		\begin{tabular}{ccc}
			\toprule
			Method & Graph Acc. & Conv. Acc. \\
			\midrule
			Our Approach w/o mask & 46.43\% & 15.14\% \\
			Our Approach & 60.53\% & 20.61\% \\
			\bottomrule
		\end{tabular}
		\caption{Performance of the models in Reddit-small dataset for ablation study.}
		\label{tab:ablation_mask}
	\end{table}
	
	\paragraph{Importance of the Ancestor Depth}
	
\begin{figure}[htbp]
		\includegraphics[width=\columnwidth]{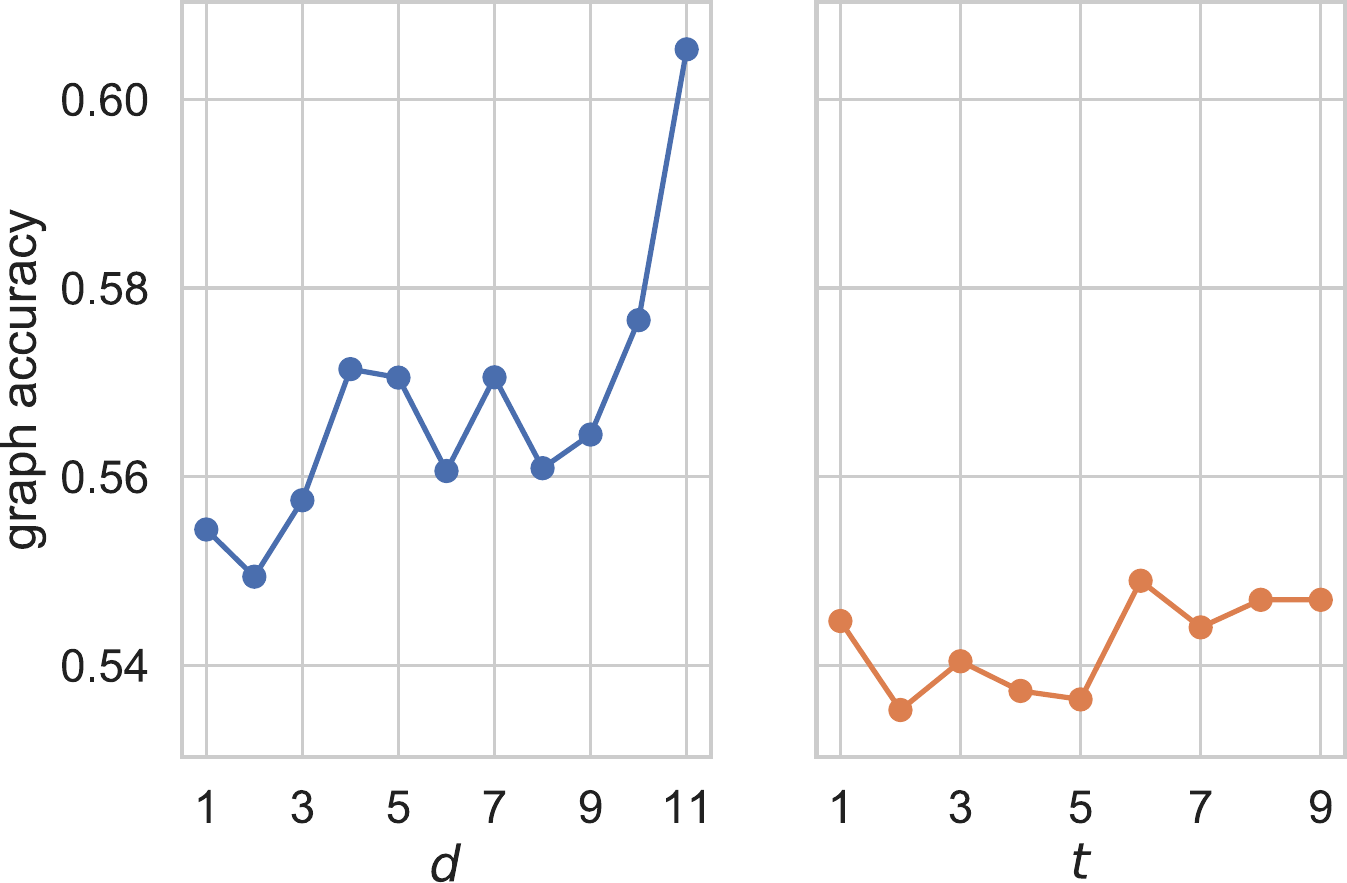}
		\caption{Performance of the model with respect to different  ancestor depth ($d$) and temporal depth ($t$).}
		\label{fig:ablation}
	\end{figure}

    Alternatively, another way of creating the mask is to use only the parent-child relation between two utterances. In other words, an utterance can only attend to its immediate parent besides the target utterance. However, since the second-stage transformer is not very deep, it may take many steps for the information of one utterance to pass to its distant descendant. To validate this hypothesis, we perform another ablation study by defining the mask in the following way. We start from the mask described in the modeling section, then we reset some cell values in the mask to zero, if the distance between two utterances in the conversation graph is greater than $d$. We call $d$ the \emph{ancestor depth} of the mask in this ablation study. There are two special cases in this model. When $d=1$, the mask is defined by the immediate parent-child `reply\_to' relation. For Reddit-small dataset, the maximum depth of the conversation tree is 12. Therefore, setting $d=11$ coincides with the mask in our model, where one non-target utterance is able to attend to all of its ancestors. We vary $d$ and report the performance of the model under different $d$ in the left plot of Figure \ref{fig:ablation}. We see generally when $d$ increases, the performance for graph accuracy increases. This result suggests including more ancestor information helps in conversation structure modeling.
	
	\paragraph{Temporal Mask}
	
	Finally, we consider another variant of mask design. The mask is designed by including temporal information only. Again, all utterances are able to attend to the target utterance. For a non-target utterance, it can attend to $t$ most recent comments before itself. We call $t$ the \emph{temporal depth} in this ablation study. The right plot in Figure \ref{fig:ablation} shows the model performances as a function of various values of $t$. There is a marginal increase when $t$ increases, but generally, the model performance is far lower than our proposed ancestor based masking approach. This suggests that the conversation structure information is more powerful than temporal information.

	\section{Conclusion}
	
	In this paper, we consider a conversation structure modeling task on Reddit and Ubuntu IRC datasets. We proposed several baseline methods as well as a novel Masked Hierarchical Transformer model, that explicitly utilizes the conversation structure. Experiments on three datasets verify that our proposed model outperforms the baseline models. The results show that taking into account the history and structure of the conversation helps in recovering the parent utterance.
	
    There are some possible directions for future work. One potential improvement for the model is to reduce the gap of the model during training and prediction since the gold conversation structure is used during training and the predicted structure is used during test time, known as `exposure bias'. Some techniques including schedule sampling \cite{bengio2015scheduled} can be applied to alleviate this problem. Also, a beam search can be applied to decode a better conversation structure instead of using a greedy way to reconstruct the conversation structure. Finally, for a better inference speed, it is possible to redesign the model so that the presentation vector of an utterance is built once. It can be done by excluding the target utterance in the mask transformer and replace the fully connected layer with a siamese network-like component, deciding if an utterance is the parent of the target utterance. 
	
	\bibliography{reference.bib}
	\bibliographystyle{aaai}
	
\end{document}